\documentclass{article}

\usepackage{arxiv}
\newcommand{\method}{METHODNAME}

\usepackage{amsmath,amsfonts,amssymb}
\usepackage{array}
\usepackage{booktabs}
\usepackage{multirow}
\usepackage{tabularx}
\usepackage{graphicx}
\usepackage[caption=false,font=normalsize,labelfont=sf,textfont=sf]{subfig}
\usepackage{textcomp}
\usepackage{stfloats}
\usepackage{placeins}
\usepackage{url}
\usepackage{verbatim}
\usepackage{cite}
\usepackage{algorithm}
\usepackage{orcidlink}
\usepackage{algpseudocode}
\usepackage{pgfplots}
\usepackage{graphicx}
\usepackage{wrapfig}
\usepackage{pgfplotstable}
\usepackage{authblk}

\usepackage{amsmath,amssymb,amsfonts,bm}
\usepackage{graphicx,booktabs,tabularx,longtable,array}
\usepackage{enumitem,microtype,xcolor,float,placeins,url}
\usepackage{hyperref}
\usepackage{orcidlink}

\hypersetup{
    colorlinks=true,
    linkcolor=blue!55!black,
    citecolor=blue!55!black,
    urlcolor=blue!55!black
}

\setlist{nosep,leftmargin=*}
\newcolumntype{Y}{>{\raggedright\arraybackslash}X}
\newcolumntype{C}[1]{>{\centering\arraybackslash}p{#1}}
\newcolumntype{P}[1]{>{\raggedright\arraybackslash}p{#1}}

\renewcommand{\method}{DG-FedReuse}
\newcommand{\norm}[1]{\left\lVert #1 \right\rVert}
\newcommand{\inner}[2]{\left\langle #1,#2 \right\rangle}

\newcommand{\saving}{modeled update-data-field uplink saving}

\usepackage{tikz}
\usepgfplotslibrary{groupplots}
\usetikzlibrary{positioning,arrows.meta,shapes.geometric,fit}
\pgfplotsset{compat=1.18}

\title{\textbf{DG-FedReuse: Proxy-Gradient-Gated Cached-Update Reuse with Matched Sparse Uplink Accounting}}

\author{
\textbf{Rahil Aftab}~\orcidlink{0009-0009-0983-4726}\\
Department of Computer Science and Engineering\\
Jamia Hamdard\\
New Delhi 110062, India\\
\texttt{rahilaftab12@gmail.com}
\and
\textbf{Vineet Kumar Rakesh}~\orcidlink{0009-0000-7102-6564}\\
Engineering Science\\
Homi Bhabha National Institute\\
Anushaktinagar, Mumbai 400094, Maharashtra, India\\
Computer and Informatics Group\\
Variable Energy Cyclotron Centre\\
1/AF, Bidhannagar, Kolkata 700064, West Bengal, India\\
\texttt{vineet@vecc.gov.in}
\and
\textbf{Soumya Mazumdar}~\orcidlink{0009-0006-3521-9557}\\
Department of Computer Science and Business Systems\\
Gargi Memorial Institute of Technology\\
Affiliated to Maulana Abul Kalam Azad University of Technology\\
Balarampur, Mouza Beralia, Baruipur, Kolkata 700144, West Bengal, India\\
\texttt{reachme@soumyamazumdar.com}
\and
\textbf{Tapas Samanta}~\orcidlink{0000-0003-0521-0747}\\
Engineering Science\\
Homi Bhabha National Institute\\
Anushaktinagar, Mumbai 400094, Maharashtra, India\\
Computer and Informatics Group\\
Variable Energy Cyclotron Centre\\
1/AF, Bidhannagar, Kolkata 700064, West Bengal, India\\
\texttt{tsamanta@vecc.gov.in}
}

\begin{document}
\maketitle

\begin{abstract}
Federated learning repeatedly incurs local optimization and model-update transmission. We study \method{}, a simulator-level mechanism that allows selected clients to contribute age-decayed cached updates when a stochastic head-gradient discrepancy proxy remains below a round-dependent threshold. A hard cache-age limit and minimum fresh-client quota constrain reuse, while fresh updates use an adaptive per-tensor Top-K numerical-field representation. Experiments cover six image-classification datasets, 50 virtual clients, Dirichlet label heterogeneity ($\alpha=0.5$), and three seeds. At a common 90-round budget, \method{} yields 83.36--85.42\% modeled update-data-field uplink saving, compared with 76.88\% for matched Top-K FedAvg; the seed-aligned accuracy differences range from $-5.29$ to $-0.14$ percentage points. Best-observed test accuracies obtained under test-controlled checkpointing are retained only as exploratory archival evidence and range from $-2.38$ to $+0.45$ percentage points relative to matched FedAvg. A symmetric dense-model-downlink sensitivity reduces the headline saving to 41.68--42.71\% and the incremental gain over Top-K FedAvg to 3.24--4.27 percentage points, demonstrating the dependence of communication conclusions on the accounting boundary. The study characterizes the proposed reuse rule in the implemented simulator; it does not establish unbiased generalization, end-to-end bandwidth reduction, runtime or energy savings, faster convergence, or superiority over existing stale-update and lazy-aggregation methods.
\end{abstract}


\section{Introduction}
Federated learning (FL) trains a shared model from decentralized data by alternating local optimization and server aggregation \cite{mcmahan2017fedavg,kairouz2021advances}. Although raw training examples remain local, each selected client ordinarily downloads a global model, performs local optimization, and uploads an update. Repeating this process can make communication and client computation important constraints, especially under heterogeneous data and partial participation.

Communication-efficient FL includes fewer communication rounds, quantization, coordinate sparsification, error-feedback mechanisms, and adaptive server optimization \cite{konecny2016strategies,reisizadeh2020fedpaq,sattler2020robust,reddi2021fedopt,li2023errorfeedback}. Most such methods reduce or transform a fresh update after local optimization. A complementary design choice is whether every selected client must recompute a fresh update in every round. Several prior methods already exploit memorized, stale, lazily transmitted, or recycled information \cite{gu2021mifa,richtarik2022threepc,rodio2024fedstale,kim2025fedluar}. The relevant novelty question is therefore not whether update reuse exists, but whether the particular gate, safeguards, sparse-update accounting, and empirical behavior studied here add a distinct and useful mechanism.

\method{} uses a stochastic proxy-gradient discrepancy to choose between a fresh local-training path and a cached-update path. A round-dependent threshold, a hard cache-age limit, and a minimum fresh-client quota constrain reuse. Fresh updates refresh client-specific state and pass through an adaptive Top-K numerical-field representation; reused updates are attenuated according to cache age. The proxy is not treated as a direct measurement of concept drift because it also reflects mini-batch sampling, data augmentation, and model-state effects.

The paper asks the following bounded question:
\begin{quote}
\emph{At a fixed communication-round budget, when the same sparse-update rule and coordinate-retention ratio are applied to FedAvg-family controls, how much additional modeled update-data-field uplink saving is associated with the implemented cached-update reuse rule, and what accuracy differences accompany it?}
\end{quote}

The contribution is an implementation and evidence study rather than a convergence or networking claim. Specifically, the paper provides:
\begin{itemize}
  \item an implementation-faithful specification of client-indexed cached-update reuse with a stochastic proxy-gradient gate, age forcing, a minimum fresh-client quota, staleness decay, and sample-weighted aggregation;
  \item deterministic numerical-field accounting for an adaptive dense/value--index/bitmap Top-K representation, with explicit separation of update uplink from excluded model downlink and protocol traffic;
  \item a six-dataset, three-seed matched sparse-control study whose primary summary uses a common 90-round budget, plus clearly labelled archival test-controlled results and mechanism-oriented secondary analyses; and
  \item an explicit audit of claim boundaries, including missing closest-method baselines, test-controlled selection, stochastic gate reliability, error feedback, complete communication measurement, and convergence analysis.
\end{itemize}

\section{Related work and positioning}
\subsection{Federated optimization and partial participation}
FedAvg established the local-SGD and sample-weighted aggregation template \cite{mcmahan2017fedavg}. FedProx adds a proximal term to mitigate client drift under heterogeneous data and systems \cite{li2020fedprox}; SCAFFOLD instead uses client and server control variates \cite{karimireddy2020scaffold}. FedOpt applies adaptive server optimization, including FedAdam-like moment updates \cite{reddi2021fedopt}. These methods primarily modify optimization or aggregation rather than deciding whether a selected client should execute a fresh local-training path.

Partial participation motivates server memory. FedVARP stores client-indexed information to reduce variance from partial participation \cite{jhunjhunwala2022fedvarp}. MIFA memorizes the latest updates from unavailable devices and uses them to correct participation-related bias \cite{gu2021mifa}. FedStale combines fresh and stale updates through a tunable interpolation and analyzes how participation and data heterogeneity affect stale-update utility \cite{rodio2024fedstale}. These methods are especially close because they establish that client-indexed historical updates can be algorithmically useful, although their participation models and update rules differ from \method{}.

\subsection{Compression, error feedback, and bidirectional accounting}
Quantization and sparsification reduce the representation size of fresh updates \cite{konecny2016strategies,reisizadeh2020fedpaq,sattler2020robust,stich2018sparsified}. Direct Top-K is a biased compressor. Error feedback can compensate for discarded information, and its behavior under partial participation requires specific analysis \cite{li2023errorfeedback}. The present implementation does not use error feedback, so matched Top-K controls isolate the implemented codec but do not represent the strongest known sparse baseline.

Most reported communication metrics in this study concern update uplink. Downlink can be a distinct systems bottleneck, and methods such as DoCoFL explicitly target model broadcast compression \cite{dorfman2023docofl}. Accordingly, we report a simple dense-downlink sensitivity in addition to the original uplink-only accounting, but do not substitute that sensitivity for measured bidirectional traffic.

\subsection{Lazy aggregation and update recycling}
Lazy or recycled-gradient methods form the closest conceptual family. LAQ suppresses quantized gradient transmission based on innovation \cite{sun2022laq}. LBGM exploits a low-rank gradient subspace and recycles update information through compact coefficients \cite{azam2022recycling}. The 3PC framework provides a general compressor class and a theory connecting lazy aggregation and error feedback \cite{richtarik2022threepc}. GradSkip studies conditional local computation and communication in distributed optimization \cite{maranjyan2025gradskip}. FedLUAR recycles selected layer updates at the server \cite{kim2025fedluar}. These results rule out a broad first-use or first-recycling claim.

Table~\ref{tab:closest} positions mechanisms rather than reporting an empirical ranking. The closest methods were not rerun under the present client partitions, models, stopping rules, and accounting boundary; their absence from the experimental baseline set is a central limitation.

\begin{table}[H]
\centering
\caption{Mechanism-level positioning. ``Not evaluated'' means that no head-to-head performance claim is made.}
\label{tab:closest}
\scriptsize
\begin{tabularx}{\textwidth}{P{0.15\textwidth}P{0.21\textwidth}YP{0.22\textwidth}}
\toprule
Method & Historical information used & Main decision structure & Status in this study \\
\midrule
MIFA \cite{gu2021mifa} & Latest updates from unavailable clients & Memory-augmented aggregation under device unavailability & Closest client-memory method; not evaluated \\
FedStale \cite{rodio2024fedstale} & Stale non-participant updates & Convex combination of fresh and stale contributions & Closest stale-update method; not evaluated \\
3PC \cite{richtarik2022threepc} & Evolving compressed reference points & General compressor framework for lazy aggregation & Closest theoretical framework; not evaluated \\
LAQ \cite{sun2022laq} & Quantized gradient history & Innovation-triggered lazy transmission & Related mechanism; not evaluated \\
LBGM \cite{azam2022recycling} & Low-rank gradient information & Coefficient-based gradient reconstruction and recycling & Related mechanism; not evaluated \\
GradSkip \cite{maranjyan2025gradskip} & Method state supporting skip decisions & Conditional local computation and communication & Related mechanism; not evaluated \\
FedLUAR \cite{kim2025fedluar} & Previous layer updates & Layer-wise server-side update recycling & Closest recent recycling method; not evaluated \\
\method{} & Client-indexed sparse full-model update & Stochastic head-proxy gate, age cap, fresh quota, and decay & Evaluated mechanism \\
\bottomrule
\end{tabularx}
\end{table}

\subsection{Role of the evaluated controls}
FedAvg and FedProx are mechanism controls because they preserve the FedAvg-family aggregation structure and use the same Top-K rule. FedProx uses the same proximal-objective form as the fresh path of \method{}, but not the same coefficient. The separately tuned FedAdam suite checks a different server optimizer. These controls help interpret the implemented mechanism, but they are not substitutes for MIFA, FedStale, 3PC-derived lazy aggregation, FedLUAR, or Top-K with error feedback. Consequently, the paper does not claim superiority over the stale-update, lazy-aggregation, or compression literature.

\section{Problem formulation and claim boundary}
Let $N$ clients hold local objectives $F_i(w)$ with nonnegative sample weights $p_i$ satisfying $\sum_i p_i=1$. The global objective is
\begin{equation}
  \min_w F(w), \qquad F(w)=\sum_{i=1}^{N} p_i F_i(w).
  \label{eq:objective}
\end{equation}
At round $t$, the server samples $m$ clients without replacement, forming $\mathcal{S}_t$. A conventional selected client starts from $w_t$, performs $E$ local epochs, and returns $\Delta_{i,t}=w_{i,t}^{(E)}-w_t$. \method{} partitions $\mathcal{S}_t$ into an active set $\mathcal{A}_t$, which computes fresh updates, and a reuse set $\mathcal{R}_t=\mathcal{S}_t\setminus\mathcal{A}_t$, which contributes client-indexed cached updates.

\begin{figure}[t]
  \centering
  \includegraphics[width=\textwidth]{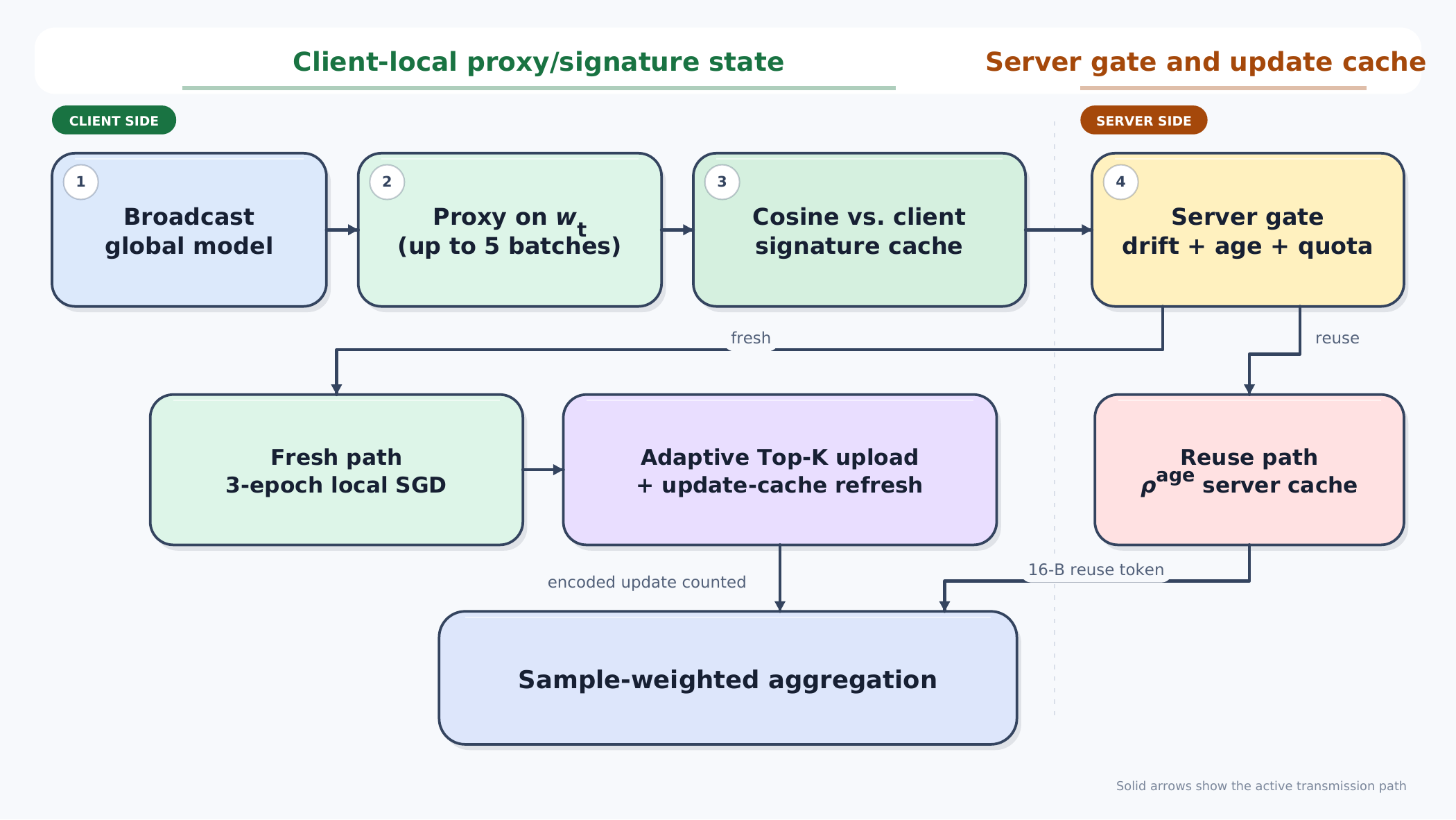}
  \caption{Logical interpretation of the single-process mechanism. A client computes a stochastic proxy-gradient discrepancy and returns a rankable scalar score; the server retains cached model updates and applies age, quota, and decay rules. Only numerical fields for fresh model updates and a conventional 16-byte reused-event charge enter the primary counter. Model downlink, representation metadata, transport framing, and most control traffic are excluded.}
  \label{fig:method}
\end{figure}

The evaluated implementation is a single-process simulator with explicit client objects and server state, not a distributed network deployment. The server invokes proxy routines and reads cached state in process. Figure~\ref{fig:method} gives one logical mapping consistent with the quota rule: a client retains its signature and returns a scalar discrepancy score, while the server retains the cached model update, ranks scores when quota promotion is required, and applies age and reuse rules. A design that transmits proxy vectors or stores update caches at clients would have different communication and storage costs.

The primary communication outcome is therefore a numerical-field ratio for client-to-server model updates, not measured bandwidth. The metric excludes model broadcast, tensor identifiers, representation-mode identifiers, framing, acknowledgements, retransmissions, and most gate/control information. A separate sensitivity adds one dense model download per selected client per round, but still does not represent a serialized protocol.

\section{DG-FedReuse}
\subsection{Stochastic proxy-gradient discrepancy}
For a selected client with a valid cache, the implementation loads the round-start global state $w_t$ and accumulates gradients for parameters whose names contain \texttt{classifier} or \texttt{fc} over up to five shuffled proxy mini-batches. The model remains in training mode; CIFAR clients can therefore include random augmentation and batch-normalization state behavior. Let $p_{i,t}$ denote the resulting flattened proxy and let $h_i$ denote the cached signature. The gate uses clipped cosine distance
\begin{equation}
  D_{i,t}=1-\operatorname{clip}\!\left(
  \frac{\inner{p_{i,t}}{h_i}}{\norm{p_{i,t}}_2\norm{h_i}_2+10^{-12}},-1,1
  \right).
  \label{eq:drift}
\end{equation}
A cacheless client receives $D_{i,t}=1$ and is activated. Because $D_{i,t}$ depends on the global state, sampled proxy batches, augmentation, and model-mode effects, we call it a stochastic proxy-gradient discrepancy rather than a direct measurement of distributional or concept drift. Its repeatability was not independently evaluated.

\subsection{Threshold, age, and forced freshness}
The round-dependent threshold is
\begin{equation}
  \tau_t=\max\!\left(\tau_{\min},\tau_0 e^{-\gamma t}\right).
  \label{eq:threshold}
\end{equation}
Because the active condition is $D_{i,t}\geq\tau_t$, decreasing $\tau_t$ makes fresh training easier to trigger. If $a_{i,t}=t-r_i$ is cache age and $S$ is maximum staleness, the preliminary active set contains every selected client satisfying
\begin{equation}
  \text{cacheless} \quad\lor\quad D_{i,t}\geq\tau_t
  \quad\lor\quad a_{i,t}\geq S.
  \label{eq:gate}
\end{equation}
If fewer than $\lceil qm\rceil$ clients satisfy Eq.~\eqref{eq:gate}, the server promotes non-active selected clients in decreasing $D_{i,t}$ order until the quota is met. The server therefore requires a rankable scalar score, not only a binary eligibility flag.

\subsection{Fresh and reused paths}
An active client performs $E$ local epochs of SGD. The \method{} fresh path uses the proximal-objective form
\begin{equation}
  F_i(w)+\frac{\mu}{2}\norm{w-w_t}_2^2.
  \label{eq:prox}
\end{equation}
The trained-model delta is sparsified, represented as dense-shaped sparse tensors in the current cache implementation, and stored in client-indexed server state. Separately, the implementation draws another proxy mini-batch sequence and recomputes a head-gradient signature at the same round-start global state $w_t$. Denoting this stochastic refresh proxy by $p_{i,t}^{\mathrm{ref}}$, the signature cache is updated by
\begin{equation}
  h_i \leftarrow 0.9h_i+0.1p_{i,t}^{\mathrm{ref}},
  \label{eq:ema}
\end{equation}
with direct initialization when no signature exists. Thus, a cacheful active event computes both a gating proxy and a separately sampled refresh proxy.

A non-active selected client contributes
\begin{equation}
  \widetilde{\Delta}_{i,t}=\rho^{a_{i,t}}\widehat{\Delta}_i,
  \label{eq:reuse}
\end{equation}
where $\widehat{\Delta}_i$ is the cached sparse update and $0<\rho<1$. Both fresh and reused contributions retain the selected client's sample weight. With $n_i$ local examples, the round update is
\begin{equation}
  \overline{\Delta}_t=\sum_{i\in\mathcal{S}_t}
  \frac{n_i}{\sum_{j\in\mathcal{S}_t}n_j}
  \begin{cases}
    \widehat{\Delta}_{i,t}, & i\in\mathcal{A}_t,\\
    \rho^{a_{i,t}}\widehat{\Delta}_i, & i\in\mathcal{R}_t.
  \end{cases}
  \label{eq:aggregate}
\end{equation}
The primary FedAvg-family server applies $w_{t+1}=w_t+\eta_s\overline{\Delta}_t$ with $\eta_s=1$.

\subsection{Adaptive Top-K numerical-field model}
For each tensor with $d$ coordinates and element width $b$ bytes, magnitude Top-K retains
\begin{equation}
  k=\min\{d,\max(1,\lfloor rd\rfloor)\}
  \label{eq:k}
\end{equation}
coordinates at ratio $r$. The accounting code selects the smallest numerical-field size among
\begin{align}
  B_{\mathrm{dense}} &= db, \\
  B_{\mathrm{pairs}} &= k(b+4), \\
  B_{\mathrm{bitmap}} &= kb+\left\lceil\frac{d}{8}\right\rceil.
  \label{eq:codec}
\end{align}
The 4-byte term represents an int32 coordinate index. These expressions count tensor values, indices, and bitmap bits. They do not include tensor identity, dimensions, dtype, selected-mode identifiers, object serialization, alignment, checksums, or transport framing. The current compressor also discards unselected coordinates without error feedback \cite{li2023errorfeedback}.

\subsection{Communication accounting}
Let $B_{i,t}$ be the sum of Eq.~\eqref{eq:codec}'s selected numerical fields over tensors for a fresh client, and let $B_0$ be the dense state-update size. The simulator charges 16 bytes for each reused-client event. This is a fixed convention, not a measured wire representation. It records
\begin{align}
  M_t &= \sum_{i\in\mathcal{A}_t}B_{i,t}+16|\mathcal{R}_t|,\\
  M_t^{\mathrm{dense}} &= mB_0,\\
  s_t &= 1-\frac{M_t}{M_t^{\mathrm{dense}}}.
  \label{eq:saving}
\end{align}
Run summaries average $s_t$, whereas the fixed-round table uses cumulative bytes through round 90. The phrase ``\saving{}'' always refers to this restricted boundary.

For a transparent sensitivity, suppose each selected client also receives one dense model of size $B_0$ in every round and compare against a dense bidirectional reference of $2mB_0$. Then
\begin{equation}
  s_t^{\mathrm{sym}}=1-\frac{mB_0+M_t}{2mB_0}=\frac{s_t}{2}.
  \label{eq:symmetric}
\end{equation}
Equation~\eqref{eq:symmetric} is not a measured system result; it only illustrates how an uncompressed downlink changes the ratio.

\begin{table}[H]
\centering
\caption{Logical events and treatment in the reported accounting. No network serialization was executed.}
\label{tab:accounting}
\small
\begin{tabularx}{\textwidth}{P{0.24\textwidth}P{0.27\textwidth}P{0.13\textwidth}Y}
\toprule
Event & Logical placement/direction & Counted? & Treatment \\
\midrule
Global-model broadcast & Server $\rightarrow$ selected client & No in $s_t$ & Added only in the symmetric sensitivity of Eq.~\eqref{eq:symmetric} \\
Proxy computation/signature & Client-local & No & No proxy vector is serialized in the logical mapping \\
Scalar discrepancy and gate control & Client $\leftrightarrow$ server & No & Rankable score and instructions are outside the counter \\
Fresh model-update fields & Active client $\rightarrow$ server & Yes & Values, int32 indices, or bitmap fields from Eq.~\eqref{eq:codec} \\
Reused cached contribution & Server-resident update cache & Yes, by convention & Fixed 16-byte event charge \\
Representation and transport metadata & Protocol/transport layer & No & Tensor schema, framing, acknowledgement, and retransmission are not simulated \\
\bottomrule
\end{tabularx}
\end{table}

\subsection{Operational invariants}
The implementation supplies safeguards, not a convergence theorem. After quota promotion, $|\mathcal{A}_t|\geq\lceil qm\rceil$. A selected client can be reused only when $a_{i,t}<S$; Eq.~\eqref{eq:gate} forces a refresh at or beyond the age limit. Because $0<\rho<1$,
\begin{equation}
  \norm{\widetilde{\Delta}_{i,t}}_2
  =\rho^{a_{i,t}}\norm{\widehat{\Delta}_i}_2
  \leq \norm{\widehat{\Delta}_i}_2.
  \label{eq:reuse-bound}
\end{equation}
This magnitude bound does not guarantee alignment with the current descent direction.

\begin{center}
\fbox{\begin{minipage}{0.94\textwidth}
\textbf{Algorithmic round $t$ of DG-FedReuse}\par\smallskip
\begin{enumerate}[label=\arabic*.]
  \item Sample $m$ clients and provide $w_t$.
  \item For each cacheful selected client, compute the stochastic proxy-gradient discrepancy $D_{i,t}$; activate cacheless, threshold-triggered, or age-forced clients.
  \item Promote the largest remaining discrepancy scores until at least $\lceil qm\rceil$ clients are active.
  \item Each active client performs proximal local SGD, applies Top-K, contributes the fresh sparse update, and refreshes update and signature caches.
  \item Each reusable client contributes its cached update scaled by $\rho^{a_{i,t}}$.
  \item Aggregate all selected-client contributions with Eq.~\eqref{eq:aggregate}, record the restricted byte counter, and execute the configured evaluation and stopping logic.
\end{enumerate}
\end{minipage}}
\end{center}
\noindent\emph{The procedure omits unexecuted networking, secure aggregation, privacy, and adversarial-robustness layers.}

\section{Experimental methodology}
\subsection{Datasets, partitioning, and models}
We use six image-classification tasks (Table~\ref{tab:data}). The internal key \texttt{femnist} loads EMNIST Balanced; it is not a writer-partitioned federated character dataset. Dirichlet allocation with $\alpha=0.5$ partitions training labels over 50 virtual clients. MNIST, FashionMNIST, EMNIST Balanced, and PathMNIST use a shallow CNN; CIFAR-10 and CIFAR-100 use torchvision ResNet-18 \cite{he2016resnet}. CIFAR training uses random cropping and horizontal flipping.

PathMNIST provides train, validation, and test partitions, but the historical final configurations set \texttt{evaluation\_split: test}. The other final tasks likewise use their canonical test sets for periodic evaluation. Consequently, final-test observations were used for early stopping and best-checkpoint selection; this design creates selection bias and prevents the reported maxima from being interpreted as unbiased held-out estimates \cite{cawley2010selectionbias}.

\begin{table}[t]
\centering
\caption{Executed dataset and model scope. Counts are the train/test examples used by the loaders.}
\label{tab:data}
\small
\begin{tabular}{lrrrl}
\toprule
Dataset & Classes & Train & Test & Model \\
\midrule
MNIST & 10 & 60,000 & 10,000 & CNN \\
FashionMNIST \cite{xiao2017fashion} & 10 & 60,000 & 10,000 & CNN \\
EMNIST Balanced \cite{cohen2017emnist} & 47 & 112,800 & 18,800 & CNN \\
PathMNIST v2 \cite{yang2023medmnist} & 9 & 89,996 & 7,180 & CNN \\
CIFAR-10 \cite{krizhevsky2009cifar} & 10 & 50,000 & 10,000 & ResNet-18 \\
CIFAR-100 \cite{krizhevsky2009cifar} & 100 & 50,000 & 10,000 & ResNet-18 \\
\bottomrule
\end{tabular}
\end{table}

\subsection{Executed protocol}
Each primary run uses 50 clients, 10 selected per round, three local epochs, batch size 128, SGD learning rate 0.05, momentum 0.9, weight decay $10^{-4}$, and server learning rate 1.0. Proxy batch size is 64. The maximum is 1,000 rounds; evaluation occurs at round 1 and every five rounds thereafter. Early stopping has patience eight evaluation events. Mixed precision is enabled on CUDA, while Top-K preparation uses a CPU path. Seeds are 101, 202, and 303.

The code seeds Python, NumPy, and PyTorch, but enables cuDNN benchmarking and does not enforce deterministic algorithms. Moreover, different methods consume random numbers through different execution paths. Thus, equal seed labels do not guarantee bitwise determinism or identical client-selection and augmentation sequences across methods; reported within-seed differences are described as seed-aligned rather than fully paired experimental blocks.

For \method{}, the frozen gate is $\tau_0=0.9116988$, $\tau_{\min}=0.8134931$, $\gamma=0.01$, $S=4$, $\rho=0.7279635$, $q=0.30$, and $\mu=3.4156\times10^{-4}$. All primary controls use $r=0.2$. FedProx uses $\mu=5\times10^{-4}$; it therefore shares the proximal form, not the coefficient, with \method{}.

\subsection{Hyperparameter provenance}
Table~\ref{tab:tuning} distinguishes historical searches that used a test-labelled stream from later validation-only selection. Disjoint final seeds do not repair repeated use of the same canonical test examples for model selection \cite{cawley2010selectionbias}. The displayed final maxima and trajectories are therefore exploratory archival evidence. The common-round analysis reduces one source of bias---maximization over evaluation time---but remains post hoc and test-observed.

\begin{table}[t]
\centering
\caption{Recorded hyperparameter-selection stages. ``Dev-test'' denotes use of a canonical test split during development.}
\label{tab:tuning}
\scriptsize
\begin{tabularx}{\textwidth}{P{0.16\textwidth}P{0.17\textwidth}P{0.13\textwidth}Y Y}
\toprule
Group & Development evidence & Budget & Search / rule & Frozen selection \\
\midrule
Shared local training & CIFAR-10 dev-test, seed 42 & 20 Optuna trials, 150 rounds & Mean best accuracy of FedAvg and DG runs; learning rate, epochs, batch, proxy batch, momentum, weight decay & $\eta=0.05$, $E=3$, batch 128, proxy 64, momentum 0.9, weight decay $10^{-4}$ \\
DG gate and reuse & CIFAR-10 dev-test, seed 42 & 20 Optuna trials, 100 rounds & Best accuracy with mean active fraction $\leq0.70$; $\tau_0,\tau_{\min},S,\rho,\mu$ & $0.9116988$, $0.8134931$, 4, $0.7279635$, $3.4156\times10^{-4}$ \\
FedProx & CIFAR-10 dev-test, seed 42 & 7-point grid, 150 rounds & $\mu\in\{0,10^{-4},5\!\times\!10^{-4},10^{-3},5\!\times\!10^{-3},10^{-2},5\!\times\!10^{-2}\}$ & $\mu=5\times10^{-4}$ \\
Top-K ratio & CIFAR-10/100 validation, seeds 41/53 & 52 initial + 32 extension runs, 150 rounds & Maximize mean saving subject to worst-dataset retention $\geq98\%$ & $r=0.2$ \\
FedAdam & CIFAR-10/100 validation, seeds 61/73 & 16 runs, 100 rounds & Server LR $\in\{0.001,0.003,0.01,0.03\}$; fixed $\beta_1=0.9,\beta_2=0.99,\tau=0.001$ & Server LR $0.01$ \\
\bottomrule
\end{tabularx}
\end{table}

The selected values were applied globally rather than tuned per dataset. Fixed design choices included $\gamma=0.01$, $q=0.30$, the head-parameter naming rule, signature EMA coefficient, five-batch proxy cap, reuse-event charge, and codec field model. They were not independently ablated.

\subsection{Top-K selection}
The initial Top-K grid covered $r=1.0,0.95,\ldots,0.40$ (52 runs). A documented post-hoc extension evaluated $r\in\{1.0,0.35,0.30,0.25,0.20,0.15,0.10,0.05\}$ (32 runs). All candidates used 150 fixed rounds and no early stopping. The rule selected the largest mean \saving{} subject to at least 98\% validation-accuracy retention relative to reuse-only $r=1.0$ on both CIFAR datasets.

\begin{table}[t]
\centering
\caption{Decision boundary from the two-stage Top-K validation study. Values below $r=0.4$ belong to the post-hoc extension.}
\label{tab:topk}
\small
\begin{tabular}{lcccc}
\toprule
$r$ & Phase & Worst retention (\%) & Mean saving (\%) & Feasible \\
\midrule
1.00 & initial reference & 100.00 & 36.47 & yes \\
0.40 & initial endpoint & 98.49 & 72.64 & yes \\
0.35 & extension & 99.95 & 75.82 & yes \\
0.30 & extension & 99.37 & 78.99 & yes \\
0.25 & extension & 98.48 & 82.16 & yes \\
\textbf{0.20} & \textbf{extension} & \textbf{98.0003} & \textbf{85.3388} & \textbf{yes} \\
0.15 & extension & 96.25 & 88.51 & no \\
0.10 & extension & 93.84 & 91.69 & no \\
0.05 & extension & 84.34 & 94.85 & no \\
\bottomrule
\end{tabular}
\end{table}

\begin{figure}[t]
  \centering
  \includegraphics[width=0.96\textwidth]{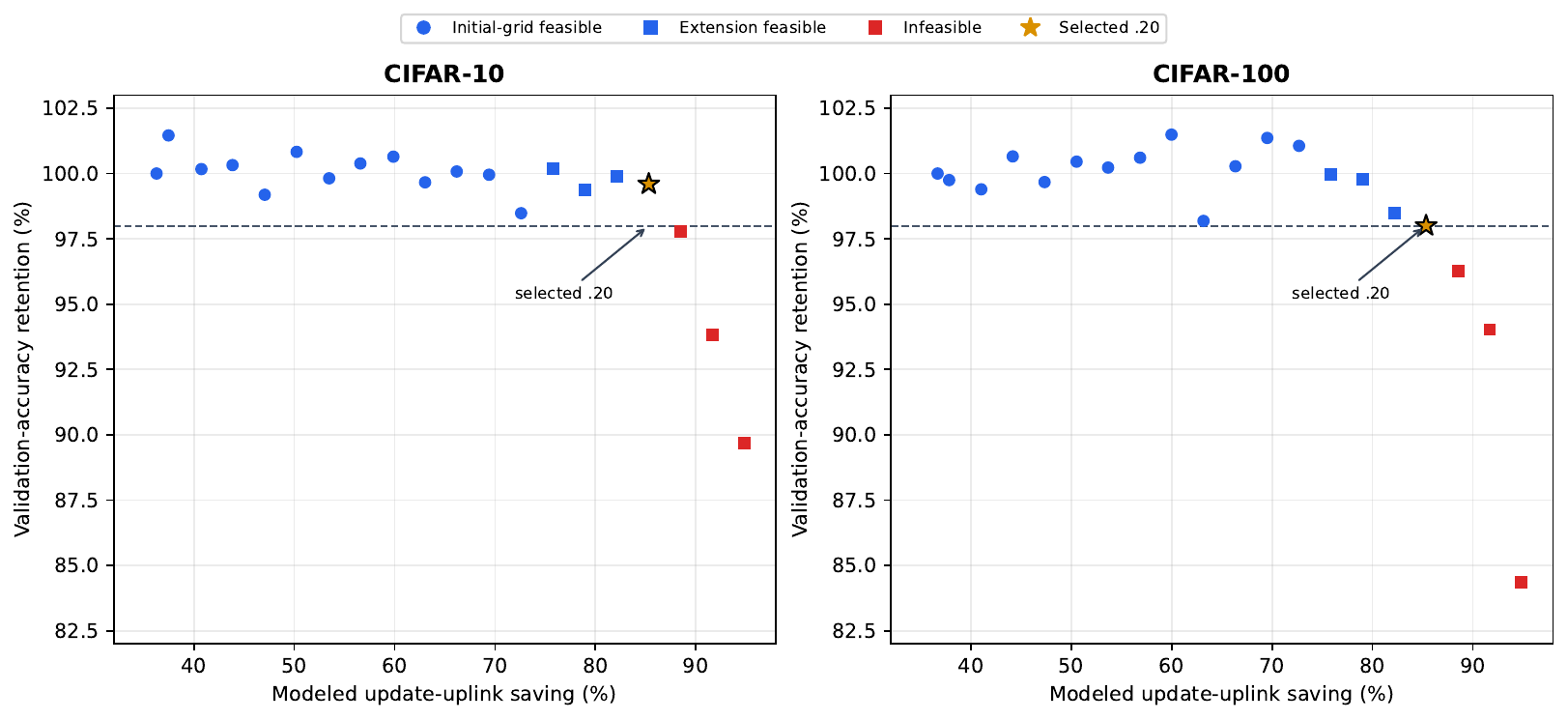}
  \caption{Validation-accuracy retention versus restricted update-uplink saving in the two-stage Top-K study. The selected $r=0.2$ candidate has 98.0003\% worst-dataset retention and therefore essentially no margin above the prespecified threshold.}
  \label{fig:topk}
\end{figure}

\subsection{Outcomes and statistical reporting}
The primary descriptive outcome uses accuracy at round 90, the earliest terminal round among all 54 FedAvg/FedProx/\method{} runs, and cumulative restricted bytes through the same round. This fixes the communication-round budget and avoids maximizing accuracy over evaluation times. The historical maximum test accuracy before stopping is reported separately as archival evidence.

Tables report arithmetic mean $\pm$ sample standard deviation over three seeds. With $n=3$, no null-hypothesis significance test, confidence claim, or equivalence claim is made. Seed-aligned differences use the same seed labels but may not share identical client schedules or augmentation draws. The primary archive contains $6\times3\times3=54$ publication-table runs. The reuse-only suite contains 36 FedAvg/\method{} runs, and the FedAdam final suite contains 18 runs after a 16-run validation search.

\section{Results}
\subsection{Fixed-round matched sparse-control comparison}
Table~\ref{tab:round90} is the primary descriptive comparison. Round 90 is the earliest terminal round among the 54 primary runs, so every trajectory contains the evaluation and no missing value is imputed. Matched Top-K FedAvg records 76.88\% cumulative restricted uplink saving. \method{} records 83.36--85.42\%, an additional 6.48--8.54 percentage points under the same field model. Accuracy is lower for \method{} on every dataset at this budget, with seed-aligned mean differences from $-0.14$ to $-5.29$ percentage points.

\begin{table}[H]
\centering
\caption{Primary fixed-budget comparison at round 90. Accuracy and cumulative restricted update-uplink saving are mean $\pm$ sample SD over seeds 101/202/303. Differences are seed-aligned, not guaranteed to use identical random trajectories.}
\label{tab:round90}
\scriptsize
\resizebox{\textwidth}{!}{%
\begin{tabular}{lccccc}
\toprule
Dataset & FedAvg acc. (\%) & DG acc. (\%) & DG--FedAvg (pp) & FedAvg save (\%) & DG save (\%) \\
\midrule
MNIST & $98.93\pm0.07$ & $98.73\pm0.27$ & $-0.20\pm0.21$ & $76.88\pm0.00$ & $83.36\pm0.24$ \\
FashionMNIST & $88.66\pm0.25$ & $88.23\pm1.28$ & $-0.43\pm1.53$ & $76.88\pm0.00$ & $83.76\pm0.49$ \\
EMNIST Balanced & $86.21\pm0.81$ & $86.06\pm0.97$ & $-0.14\pm0.37$ & $76.88\pm0.00$ & $84.69\pm0.12$ \\
PathMNIST & $83.77\pm1.72$ & $80.98\pm1.85$ & $-2.78\pm2.88$ & $76.88\pm0.00$ & $84.46\pm0.36$ \\
CIFAR-10 & $67.50\pm1.80$ & $62.21\pm1.66$ & $-5.29\pm1.18$ & $76.88\pm0.00$ & $85.36\pm0.12$ \\
CIFAR-100 & $38.36\pm0.61$ & $36.04\pm0.18$ & $-2.31\pm0.45$ & $76.88\pm0.00$ & $85.42\pm0.08$ \\
\bottomrule
\end{tabular}}
\end{table}

The fixed-budget evidence therefore shows a communication--accuracy trade-off, not utility preservation. The deficit is small on MNIST, FashionMNIST, and EMNIST Balanced at round 90, but is substantial on PathMNIST and both CIFAR tasks. No result establishes lower bytes to a common target accuracy because target attainment was not prespecified and several trajectories have different convergence rates.

\subsection{Dense-downlink sensitivity}
Equation~\eqref{eq:symmetric} adds one uncompressed model download per selected client per round. Under this deliberately simple symmetric reference, Top-K FedAvg's saving becomes 38.44\%, and \method{} becomes 41.68--42.71\% (Table~\ref{tab:symmetric}). The incremental advantage is then 3.24--4.27 percentage points. Control traffic and representation metadata remain excluded, so these values are still model-based rather than measured.

\begin{table}[H]
\centering
\caption{Round-90 communication sensitivity with one dense model downlink plus the modeled uplink, relative to dense downlink plus dense uplink.}
\label{tab:symmetric}
\small
\begin{tabular}{lccc}
\toprule
Dataset & Top-K FedAvg saving (\%) & DG saving (\%) & DG gain (pp) \\
\midrule
MNIST & 38.44 & 41.68 & 3.24 \\
FashionMNIST & 38.44 & 41.88 & 3.44 \\
EMNIST Balanced & 38.44 & 42.35 & 3.91 \\
PathMNIST & 38.44 & 42.23 & 3.79 \\
CIFAR-10 & 38.44 & 42.68 & 4.24 \\
CIFAR-100 & 38.44 & 42.71 & 4.27 \\
\bottomrule
\end{tabular}
\end{table}

The difference between Tables~\ref{tab:round90} and \ref{tab:symmetric} is consequential. A headline based only on update uplink overstates the fraction of a bidirectional dense reference removed when model broadcast remains uncompressed. This sensitivity is consistent with the broader observation that downlink requires separate treatment in cross-device FL \cite{dorfman2023docofl}.

\subsection{Archival test-controlled checkpoint summary}
Table~\ref{tab:primary} preserves the originally reported best-observed test results for auditability. Because the test stream controlled checkpoint selection and early stopping, the values are not unbiased final-test estimates. They should not be used to assert equivalence, superiority, or generalization.

\begin{table}[H]
\centering
\caption{Exploratory archival summary under test-controlled checkpointing. All methods use $r=0.2$ and seeds 101/202/303. Saving is the run mean of the restricted update-uplink counter.}
\label{tab:primary}
\scriptsize
\resizebox{\textwidth}{!}{%
\begin{tabular}{lcccccccc}
\toprule
Dataset & FedAvg acc. & FedProx acc. & DG acc. & DG--FedAvg & FedAvg save & FedProx save & DG save & DG gain \\
 & (\%) & (\%) & (\%) & (pp) & (\%) & (\%) & (\%) & (pp) \\
\midrule
MNIST & $99.07\pm0.09$ & $99.11\pm0.07$ & $99.13\pm0.07$ & $+0.06$ & $76.88\pm0.00$ & $76.88\pm0.00$ & $84.32\pm0.32$ & $+7.44$ \\
FashionMNIST & $90.79\pm0.26$ & $90.73\pm0.41$ & $91.23\pm0.05$ & $+0.44$ & $76.88\pm0.00$ & $76.88\pm0.00$ & $84.67\pm0.26$ & $+7.79$ \\
EMNIST Balanced & $87.37\pm0.19$ & $87.44\pm0.22$ & $87.24\pm0.26$ & $-0.13$ & $76.88\pm0.00$ & $76.88\pm0.00$ & $85.05\pm0.13$ & $+8.17$ \\
PathMNIST & $85.07\pm1.87$ & $84.49\pm0.14$ & $82.70\pm1.35$ & $-2.38$ & $76.88\pm0.00$ & $76.88\pm0.00$ & $84.62\pm0.50$ & $+7.74$ \\
CIFAR-10 & $75.02\pm4.33$ & $75.08\pm5.16$ & $74.50\pm4.05$ & $-0.52$ & $76.88\pm0.00$ & $76.88\pm0.00$ & $85.46\pm0.15$ & $+8.58$ \\
CIFAR-100 & $45.31\pm0.54$ & $45.77\pm0.42$ & $45.76\pm1.22$ & $+0.45$ & $76.88\pm0.00$ & $76.88\pm0.00$ & $85.52\pm0.05$ & $+8.64$ \\
\bottomrule
\end{tabular}}
\end{table}

\begin{figure}[t]
  \centering
  \includegraphics[width=\textwidth]{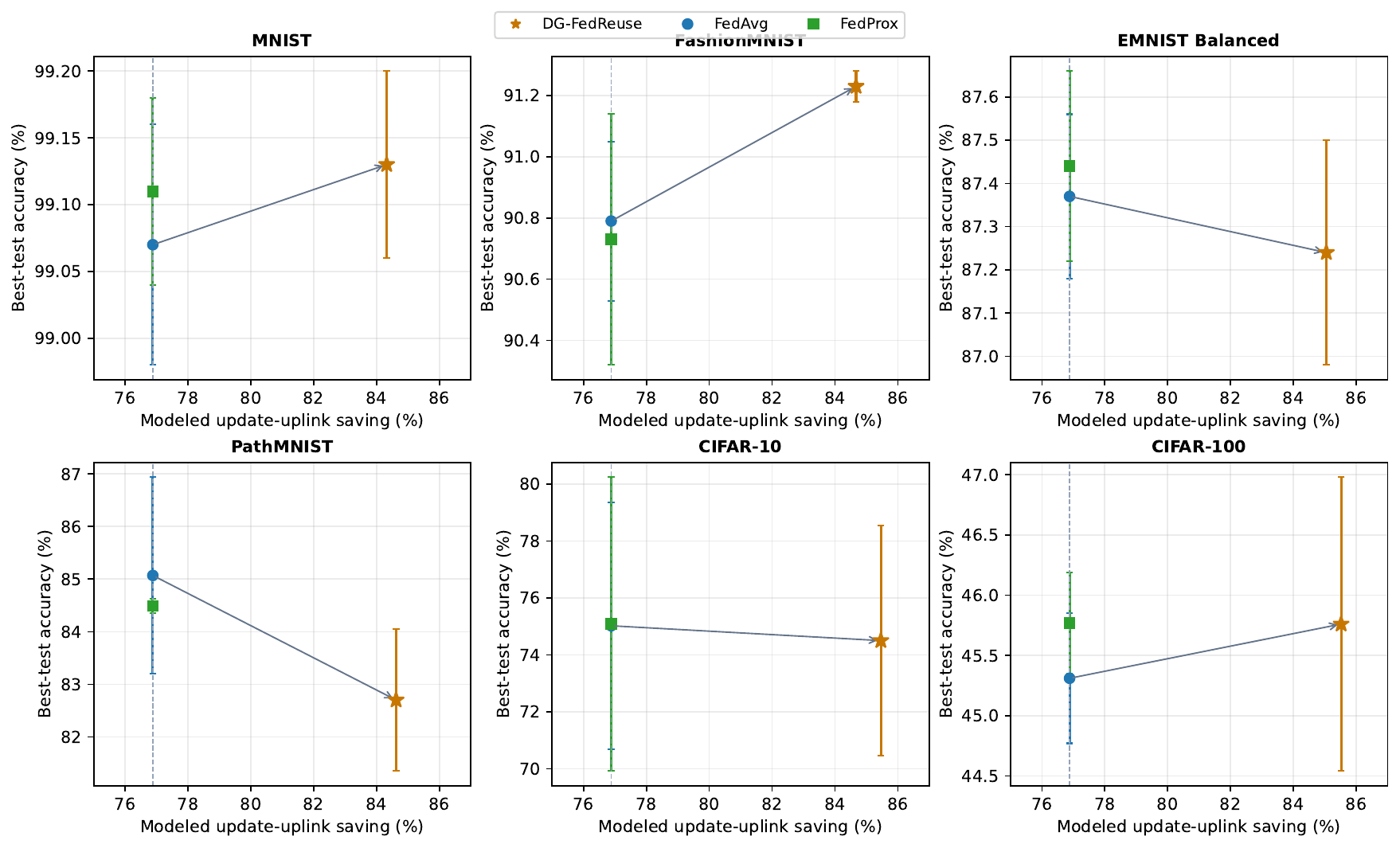}
  \caption{Exploratory archival best-observed test accuracy versus restricted update-uplink saving. Error bars are sample standard deviations over three seeds. The plot is diagnostic because checkpoint selection used the test stream.}
  \label{fig:accuracy-saving}
\end{figure}

The archival maxima are more favorable than the common-round values, particularly for CIFAR-10 and CIFAR-100. This divergence illustrates why maximizing repeatedly observed test accuracy can materially change interpretation \cite{cawley2010selectionbias}. Figure~\ref{fig:accuracy-saving} is retained as a diagnostic visualization of these archival values.

\subsection{FedAdam secondary analysis}
FedAdam was selected on CIFAR-10/CIFAR-100 validation data with development seeds 61 and 73; server learning rate 0.01, $\beta_1=0.9$, $\beta_2=0.99$, and $\tau=0.001$ were then frozen. Its final suite nevertheless uses the same test-controlled stopping design as the primary archive, so Table~\ref{tab:fedopt} is also exploratory.

\begin{table}[H]
\centering
\caption{Exploratory FedAdam summary under test-controlled checkpointing. Accuracy and rounds are mean $\pm$ sample SD over seeds 101/202/303.}
\label{tab:fedopt}
\small
\begin{tabular}{lcccc}
\toprule
Dataset & Best-observed test (\%) & Terminal test (\%) & Saving (\%) & Actual rounds \\
\midrule
MNIST & $99.07\pm0.16$ & $98.93\pm0.16$ & $76.88\pm0.00$ & $151.7\pm20.2$ \\
FashionMNIST & $91.19\pm0.25$ & $90.72\pm0.50$ & $76.88\pm0.00$ & $201.7\pm55.3$ \\
EMNIST Balanced & $87.12\pm0.23$ & $86.63\pm0.34$ & $76.88\pm0.00$ & $161.7\pm2.9$ \\
PathMNIST & $84.00\pm1.15$ & $80.97\pm1.46$ & $76.88\pm0.00$ & $150.0\pm36.1$ \\
CIFAR-10 & $76.80\pm0.32$ & $75.47\pm1.17$ & $76.88\pm0.00$ & $246.7\pm41.6$ \\
CIFAR-100 & $45.29\pm0.73$ & $44.98\pm0.59$ & $76.88\pm0.00$ & $263.3\pm53.5$ \\
\bottomrule
\end{tabular}
\end{table}

\begin{figure}[t]
  \centering
  \includegraphics[width=\textwidth]{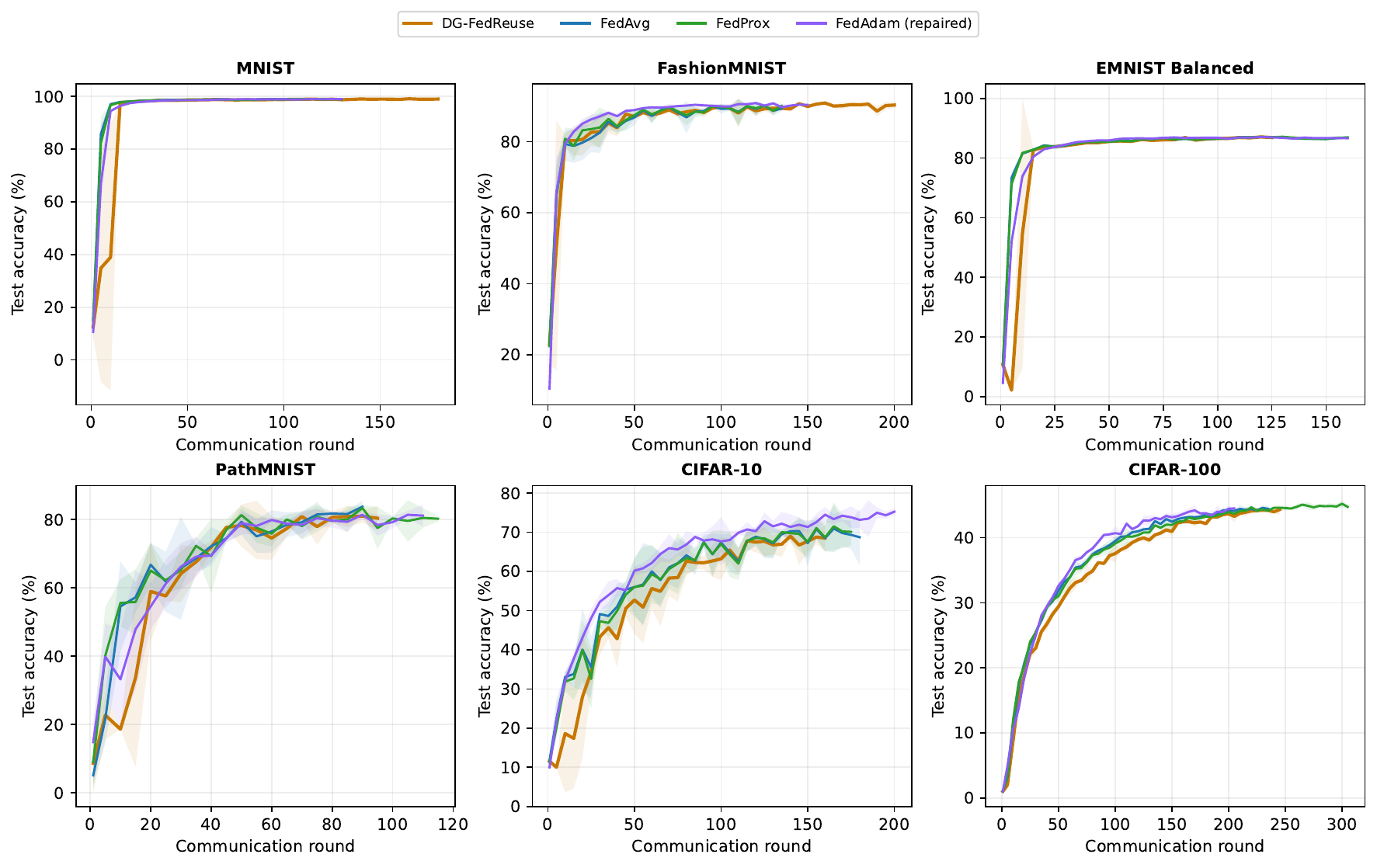}
  \caption{Test-observed trajectories for the three FedAvg-family methods and separately selected FedAdam. Lines are three-seed means and bands are sample standard deviations. Curves terminate at the earliest stopping round represented in all three corresponding runs. They are diagnostic test trajectories, not validation curves or confidence bands.}
  \label{fig:convergence}
\end{figure}

\subsection{Descriptive mechanism decomposition}
Table~\ref{tab:mechanism} juxtaposes dense FedAvg, reuse without Top-K, Top-K without reuse, and the combined mechanism. The conditions come from separately executed suites; they do not form a randomized $2\times2$ factorial experiment with identical schedules and stopping rules. The table is therefore descriptive and cannot identify an independent causal effect or interaction.

\begin{table}[H]
\centering
\caption{Descriptive cross-suite mechanism comparison. Each cell is exploratory best-observed test accuracy / restricted update-uplink saving, mean $\pm$ sample SD.}
\label{tab:mechanism}
\scriptsize
\resizebox{\textwidth}{!}{%
\begin{tabular}{lcccc}
\toprule
Dataset & Dense FedAvg & DG reuse only & $0.2$ Top-K FedAvg & DG reuse + $0.2$ Top-K \\
\midrule
MNIST & $99.10\pm0.13$ / $0$ & $99.05\pm0.07$ / $32.43\pm1.93$ & $99.07\pm0.09$ / $76.88\pm0.00$ & $99.13\pm0.07$ / $84.32\pm0.32$ \\
FashionMNIST & $90.80\pm0.04$ / $0$ & $91.14\pm0.59$ / $32.83\pm2.48$ & $90.79\pm0.26$ / $76.88\pm0.00$ & $91.23\pm0.05$ / $84.67\pm0.26$ \\
EMNIST Balanced & $87.54\pm0.07$ / $0$ & $87.52\pm0.19$ / $35.73\pm0.49$ & $87.37\pm0.19$ / $76.88\pm0.00$ & $87.24\pm0.26$ / $85.05\pm0.13$ \\
PathMNIST & $85.63\pm0.66$ / $0$ & $85.32\pm1.10$ / $34.60\pm1.10$ & $85.07\pm1.87$ / $76.88\pm0.00$ & $82.70\pm1.35$ / $84.62\pm0.50$ \\
CIFAR-10 & $79.64\pm0.69$ / $0$ & $74.38\pm2.68$ / $36.99\pm0.59$ & $75.02\pm4.33$ / $76.88\pm0.00$ & $74.50\pm4.05$ / $85.46\pm0.15$ \\
CIFAR-100 & $46.11\pm0.80$ / $0$ & $45.92\pm0.38$ / $37.28\pm0.51$ & $45.31\pm0.54$ / $76.88\pm0.00$ & $45.76\pm1.22$ / $85.52\pm0.05$ \\
\bottomrule
\end{tabular}}
\end{table}

\begin{figure}[t]
  \centering
  \includegraphics[width=\textwidth]{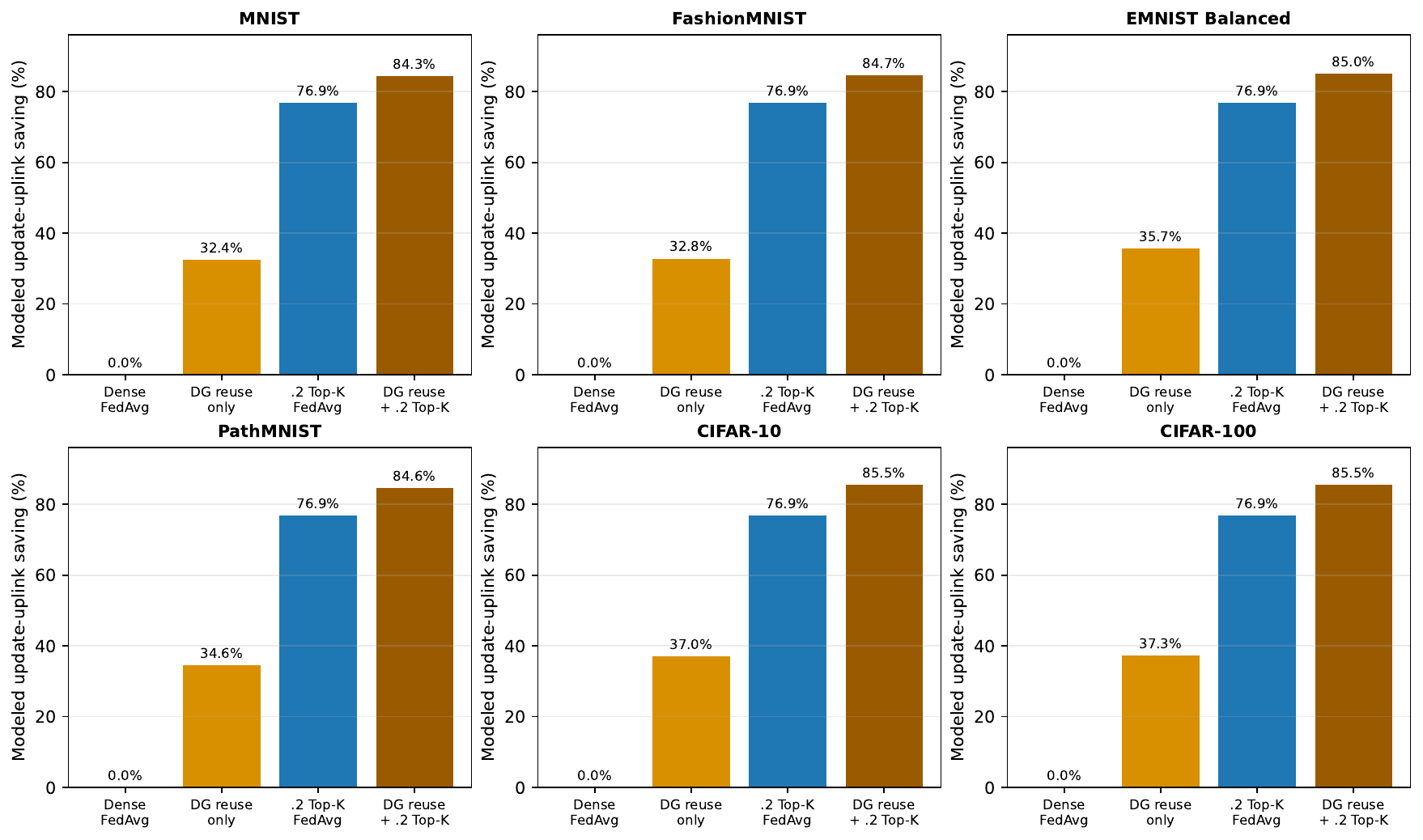}
  \caption{Restricted update-uplink saving in separately executed dense, reuse-only, Top-K-only, and combined suites. The visualization is not a factorial causal decomposition or a runtime result.}
  \label{fig:decomposition}
\end{figure}

Reuse-only saving ranges from 32.43\% to 37.28\%, which is consistent with cached-update reuse contributing additional field-count reduction beyond sparsification. However, reuse-only CIFAR-10 accuracy is 5.26 percentage points below dense FedAvg, and the separate-suite design does not estimate an interaction between reuse and Top-K.

\subsection{Evidence supported by the executed study}
The executed evidence supports the following bounded observations:
\begin{enumerate}
  \item At round 90, the implemented gate increases restricted update-uplink saving over Top-K FedAvg from 76.88\% to 83.36--85.42\%, while reducing mean accuracy by 0.14--5.29 percentage points across the six tasks.
  \item Adding one dense model downlink per selected client reduces the corresponding communication ratio to 41.68--42.71\%, with a 3.24--4.27 percentage-point incremental gain over Top-K FedAvg.
  \item Separately executed reuse-only runs produce 32.43--37.28\% restricted uplink saving, but do not constitute a factorial attribution experiment.
  \item Test-controlled maxima and trajectories are reproducible archival outputs, not unbiased held-out performance estimates.
\end{enumerate}
The study does not establish superiority over MIFA, FedStale, 3PC-derived lazy aggregation, FedLUAR, or error-feedback compression; faster convergence; lower communication to a target accuracy; reduced client FLOPs; or end-to-end network, runtime, energy, privacy, or robustness gains.

\section{Discussion}
\subsection{Interpretation of the matched sparse controls}
Applying the same $r=0.2$ field-count rule to FedAvg, FedProx, and \method{} prevents the full 76.88\% Top-K reduction from being attributed to reuse. The remaining uplink difference is associated with the implemented active/reuse decisions under the simulator's 16-byte reused-event convention. This is a useful mechanism control, but it is narrower than a state-of-the-art comparison because Top-K lacks error feedback and the closest stale-update and lazy-aggregation methods are absent.

\subsection{Accuracy--communication behavior}
The fixed-round analysis exposes slower transient progress on PathMNIST and both CIFAR tasks. The archival maxima close part of this gap only after method-dependent stopping and repeated test observation. Therefore, the method should be viewed as a tunable communication--accuracy mechanism rather than an accuracy-preserving replacement for fresh training. A clean study should choose accuracy tolerances or target levels on validation data before observing the final test set.

\subsection{What is distinctive and what is not}
Historical-update use is not new: MIFA, FedStale, 3PC-style lazy aggregation, and FedLUAR already establish memorized, stale, lazy, or recycled update mechanisms \cite{gu2021mifa,richtarik2022threepc,rodio2024fedstale,kim2025fedluar}. The distinct element evaluated here is the combination of a client-specific stochastic head-proxy score, an explicit age cap, a minimum fresh-client quota, age decay, and a matched sparse numerical-field accounting study. Whether that combination improves the accuracy--communication frontier relative to the closest methods remains unresolved.

\section{Threats to validity and limitations}
\paragraph{Test-controlled selection.}
The historical final configurations use the canonical test stream for checkpoint selection and early stopping. Early shared and gate searches also used a test-labelled development stream. The common-round table removes maximization over evaluation time but remains a post-hoc analysis of repeatedly observed test data. A confirmatory submission requires validation-derived selection and once-only final test evaluation \cite{cawley2010selectionbias}.

\paragraph{Stochastic gate validity.}
The proxy uses shuffled batches, training-mode model behavior, and random CIFAR augmentation. Its repeatability, score variance, and decision-flip rate at a fixed global/client state were not measured. Consequently, the manuscript does not interpret $D_{i,t}$ as pure client drift. A deterministic-proxy ablation and repeated-measurement reliability analysis are required to establish that the gate responds to the intended signal.

\paragraph{Compression baseline.}
Direct Top-K discards residual coordinates without error feedback. Because biased compression can alter convergence and error-feedback behavior changes under partial participation, a matched Top-K-with-error-feedback baseline is required for a stronger compression comparison \cite{li2023errorfeedback}. Client residual-buffer memory and any additional metadata must be included.

\paragraph{Modeled rather than measured communication.}
No distributed transport is executed. The primary counter excludes model downlink, tensor and mode metadata, control scores, framing, acknowledgements, and retransmissions. The symmetric sensitivity adds a dense downlink but still omits those terms. End-to-end communication requires serialized uplink and downlink measurement under an explicit protocol; downlink should be treated separately rather than assumed negligible \cite{dorfman2023docofl}.

\paragraph{Client computation and storage.}
Reusable clients still compute a proxy, and cacheful active clients compute both a gating and refresh proxy. The shallow-CNN head is large. No FLOP, latency, energy, or device-memory reduction is measured. Cached sparse updates are stored as dense-shaped CPU tensors, so server memory scales with dense model size per populated client cache.

\paragraph{Randomization and statistical power.}
Equal seed labels do not guarantee identical method-specific random trajectories, and cuDNN deterministic algorithms are not enforced. Three seeds permit descriptive standard deviations but provide weak tail and uncertainty characterization. The study therefore avoids significance and equivalence claims. A rerun should freeze client partitions and selection schedules, separate random streams by function, record environment hashes, and use more seeds with prespecified intervals or equivalence margins.

\paragraph{Mechanism attribution.}
The dense, reuse-only, Top-K-only, and combined conditions were executed in separate suites. They do not form a common $2\times2$ factorial design. Independent effects and interactions therefore cannot be claimed. A factorial experiment must use identical partitions, schedules, budgets, model-selection rules, and final evaluation.

\paragraph{Closest-method comparison.}
MIFA, FedStale, 3PC-derived lazy aggregation, FedLUAR, LAQ, LBGM, and GradSkip were not implemented under the current experimental contract. The present controls isolate an internal mechanism but cannot support a broad novelty or superiority conclusion.

\paragraph{Theory, privacy, and robustness.}
The minimum-freshness and bounded-age properties are implementation invariants, not convergence guarantees. No analysis bounds stale-direction error jointly with Top-K error. Differential privacy, secure aggregation, poisoning, gradient leakage, and adversarial robustness were not evaluated; no privacy or security guarantee is made \cite{zhu2019dlg}.

\section{Required confirmatory work}
A confirmatory evaluation should first create validation partitions exclusively from training data, freeze all hyperparameters and round/checkpoint rules, and evaluate each canonical test set once. It should then compare FedAvg, FedProx, Top-K with and without error feedback, MIFA, FedStale, a 3PC-derived lazy method, and FedLUAR under identical partitions, client schedules, budgets, and accounting. At least five independent seeds should be reported with seed-level observations, confidence intervals, and prespecified accuracy-retention or equivalence criteria.

The mechanism study should use a $2\times2$ reuse $\times$ Top-K factorial design and ablate the proxy, age cap, quota, decay, signature EMA, proxy-batch count, and threshold schedule. Gate reliability should be measured by repeated proxy evaluations at fixed states, including score variance and decision-flip rates. System validation should serialize both model downlink and update uplink, including schema, tensor and mode identifiers, control scores, framing, acknowledgements, and retransmissions; it should also report bytes and wall-clock time to a validation-defined target accuracy, proxy and training FLOPs, energy, and cache memory.

\section{Reproducibility and evidence governance}
The supplied evidence distinguishes a 54-run FedAvg/FedProx/\method{} primary archive, a 36-run dense/reuse-only suite, a complete 84-run two-stage Top-K validation study, a 16-run FedAdam validation search, and an 18-run FedAdam final archive. The larger raw matched and uncompressed archives contain additional algorithm labels and artifacts; run presence does not make every label eligible for publication claims. The manuscript tables use only the stated subsets.

Each run directory records resolved configuration, metrics, summary, log, and checkpoints where available. The compact evidence manifest reports 1,558 entries and 412 non-empty metric trajectories. The corrected release test path points to \texttt{configs/baseline\_exp/smoke\_synthetic.yaml}; the packaged repository passes 22 tests and 14 subtests. These checks establish artifact consistency, not scientific validity of test-selected estimates.

\section{Conclusion}
\method{} combines a stochastic proxy-gradient gate, bounded cache age, a minimum fresh-client quota, staleness decay, and matched Top-K numerical-field accounting. At a common 90-round budget, the implemented simulator records 83.36--85.42\% restricted update-uplink saving versus 76.88\% for Top-K FedAvg, accompanied by seed-aligned accuracy differences from $-0.14$ to $-5.29$ percentage points. Adding one dense model downlink per selected client reduces the corresponding ratio to 41.68--42.71\% and the incremental gain to 3.24--4.27 percentage points. These observations show that the mechanism changes the modeled communication--accuracy trade-off under the evaluated simulator. They do not establish unbiased final-test performance, end-to-end communication or computation savings, convergence superiority, or an advantage over existing stale-update, lazy-aggregation, recycling, and error-feedback methods. Those questions require the validation-controlled, closest-baseline, factorial, and distributed-system experiments specified above.

\appendix
\section{Exact search spaces and fixed choices}
The shared Optuna search used learning rate $\{0.001,0.005,0.01,0.05\}$, local epochs $\{1,2,3\}$, batch size $\{128,256\}$, weight decay $\{10^{-4},5\times10^{-4},10^{-3}\}$, momentum $\{0,0.9\}$, and proxy batch $\{32,64,128\}$.  The DG search sampled $\tau_0\in[0.85,0.99]$, $\tau_{\min}\in[0.40,\min(\tau_0-0.01,0.85)]$, $S\in\{2,\ldots,15\}$, $\rho\in[0.5,0.95]$, and log-uniform $\mu\in[10^{-5},10^{-2}]$.  The search constrained mean active fraction to at most 0.70.  Fixed DG choices were $\gamma=0.01$, $q=0.30$, and a head signature.  The secondary FedOpt validation fixed $\beta_1=0.9$, $\beta_2=0.99$, and stabilizer $\tau=0.001$ while searching server learning rate.

\section{Model-state and proxy sizes}
Table~\ref{tab:sizes} makes the simulator boundary concrete.  Dense update bytes use actual tensor element widths in the state dictionary.  Proxy bytes show an fp32 representation of the implemented head-gradient signature; they are not charged in Eq.~\eqref{eq:saving}.  The logical mapping in Fig.~\ref{fig:method} keeps that vector client-local.  Because the current sparse cache is stored in dense-shaped tensors, its tensor storage for $N_c$ populated client caches is exactly $N_cB_0$ before framework overhead.

\begin{table}[H]
\centering
\caption{Model, proxy, and dense-shaped update-cache sizes used to audit the accounting boundary.  GB uses $10^9$ bytes and assumes one populated update cache per client; framework overhead and signature storage are excluded from the last two columns.}
\label{tab:sizes}
\scriptsize
\resizebox{\textwidth}{!}{%
\begin{tabular}{lrrrrr}
\toprule
Dataset/model & Trainable parameters & Dense update bytes & Proxy signature bytes & Cache, 50 clients (GB) & Cache, 1,000 clients (GB) \\
\midrule
MNIST / CNN & 421,834 & 1,688,120 & 1,611,304 & 0.0844 & 1.688 \\
FashionMNIST / CNN & 421,834 & 1,688,120 & 1,611,304 & 0.0844 & 1.688 \\
EMNIST Balanced / CNN & 426,607 & 1,707,212 & 1,630,396 & 0.0854 & 1.707 \\
PathMNIST / CNN & 422,281 & 1,689,908 & 1,610,788 & 0.0845 & 1.690 \\
CIFAR-10 / ResNet-18 & 11,181,642 & 44,765,128 & 20,520 & 2.2383 & 44.765 \\
CIFAR-100 / ResNet-18 & 11,227,812 & 44,949,808 & 205,200 & 2.2475 & 44.950 \\
\bottomrule
\end{tabular}}
\end{table}

\section{Seed-aligned archival accuracy differences}
Table~\ref{tab:paired-seeds} reports the three seed-aligned observations underlying the exploratory archival DG--FedAvg best-observed test comparison. Equal seed labels use the corresponding partition protocol, but method-specific execution can consume different random-number sequences and need not produce identical client schedules. The table increases transparency but is not a paired randomized experiment, equivalence test, or significance test.

\begin{table}[H]
\centering
\caption{Seed-aligned exploratory best-observed test differences, \method{} minus Top-K FedAvg, in percentage points. The final column is mean $\pm$ sample SD across seed labels.}
\label{tab:paired-seeds}
\small
\begin{tabular}{lrrrr}
\toprule
Dataset & Seed 101 & Seed 202 & Seed 303 & Seed-aligned mean $\pm$ SD \\
\midrule
MNIST & $+0.08$ & $-0.07$ & $+0.16$ & $+0.06\pm0.12$ \\
FashionMNIST & $+0.25$ & $+0.35$ & $+0.72$ & $+0.44\pm0.25$ \\
EMNIST Balanced & $-0.14$ & $+0.10$ & $-0.35$ & $-0.13\pm0.22$ \\
PathMNIST & $+1.25$ & $-3.54$ & $-4.85$ & $-2.38\pm3.21$ \\
CIFAR-10 & $-1.71$ & $-0.15$ & $+0.30$ & $-0.52\pm1.05$ \\
CIFAR-100 & $-1.38$ & $+1.65$ & $+1.08$ & $+0.45\pm1.61$ \\
\bottomrule
\end{tabular}
\end{table}

\section{Baseline evidence disposition}
The larger archived suite contains six algorithm labels, but run completeness alone is not sufficient for scientific comparison. FedAvg and FedProx are the matched sparse mechanism controls. Historical FedOpt rows are excluded; only the separately selected 18-run FedAdam suite in Table~\ref{tab:fedopt} is summarized. SCAFFOLD would require verified control-variate evolution and complete traffic accounting, while Ditto would require personalized-client outcomes. No archived label substitutes for a correctly implemented closest-method baseline.

\section{Artifact lineage}
The numerical claims in the manuscript trace to four frozen evidence roles:
\begin{enumerate}
  \item the validation-only CIFAR Top-K study for ratio selection;
  \item the uncompressed three-seed suite for descriptive reuse-only evidence;
  \item the three-seed matched $r=0.2$ suite for the sparse-control comparison; and
  \item the validation-selected 18-run FedAdam suite for the secondary optimizer analysis.
\end{enumerate}
Resolved run configurations override template YAML and narrative documentation when they differ.  Development trials are never counted as final seeds.  A result row is admitted only after the expected dataset--method--seed coverage and required artifacts are complete.

\section*{Acknowledgments}
The authors gratefully acknowledge the support of the Variable Energy Cyclotron Centre (VECC), the Department of Atomic Energy (DAE), Government of India, for providing the infrastructure and technical environment that supported this research. The authors also thank the staff of the VECC library for their assistance during the course of this study.

\section*{Declarations}

\subsubsection*{Consent to Publish}
All authors have read and approved the final manuscript and consent to its submission and publication.

\subsubsection*{Conflict of Interest}
The authors declare that they have no known competing financial interests or personal relationships that could have appeared to influence the work reported in this paper.

\bibliographystyle{IEEEtran}
\bibliography{references}

\section*{Author Biographies}

\renewcommand{\arraystretch}{1.2}
\noindent\begin{tabular}{@{}p{0.17\textwidth} p{0.78\textwidth}@{}}

\begin{minipage}[t]{\linewidth}
\vspace{0pt}
\includegraphics[width=\linewidth]{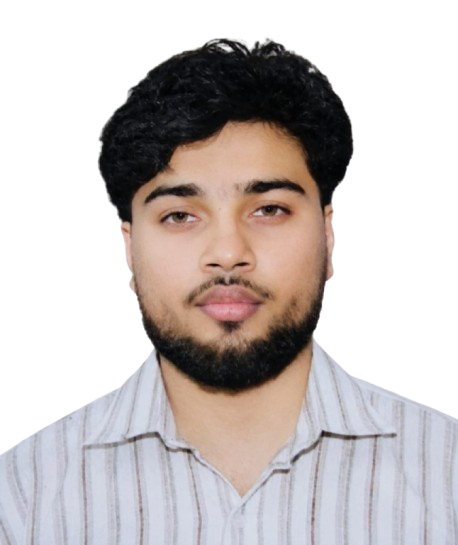}
\end{minipage}
&
\begin{minipage}[t]{\linewidth}
\vspace{0pt}
\textbf{Rahil Aftab} is an undergraduate student pursuing a Bachelor of Technology (B.Tech.) in Computer Science and Engineering at Jamia Hamdard, New Delhi, India. His research interests include artificial intelligence, computer vision, federated learning, privacy-preserving machine learning, and large language models. He has contributed to research projects involving federated learning, benchmark evaluation, and AI-driven applications. He completed a research internship at the Variable Energy Cyclotron Centre (VECC), Department of Atomic Energy, India, where he conducted research on federated learning for privacy-preserving AI systems. His broader interests include trustworthy AI and the real-world deployment of intelligent systems.
\end{minipage}
\\[1.5em]

\begin{minipage}[t]{\linewidth}
\vspace{0pt}
\includegraphics[width=\linewidth]{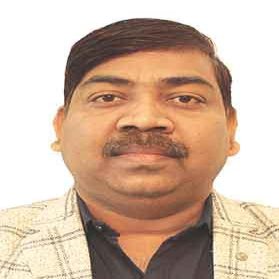}
\end{minipage}
&
\begin{minipage}[t]{\linewidth}
\vspace{0pt}
\textbf{Vineet Kumar Rakesh} is a Technical Officer (Scientific Category) at the Variable Energy Cyclotron Centre (VECC), Department of Atomic Energy, India, with over 23 years of experience in software engineering, database systems, and artificial intelligence. His research focuses on talking head generation, lip reading, and ultra-low-bitrate video compression for real-time teleconferencing. He is currently pursuing a Ph.D. at Homi Bhabha National Institute, Mumbai. Mr. Rakesh has contributed to office automation, OCR systems, and digital transformation projects at VECC. He is an Associate Member of the Institution of Engineers (India) and a recipient of the DAE Group Achievement Award.
\end{minipage}
\\[1.5em]

\begin{minipage}[t]{\linewidth}
\vspace{0pt}
\includegraphics[width=\linewidth]{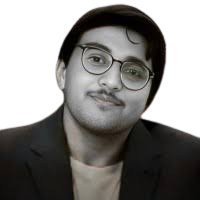}
\end{minipage}
&
\begin{minipage}[t]{\linewidth}
\vspace{0pt}
\textbf{Soumya Mazumdar} is a student researcher pursuing a B.S. in Data Science and Applications at the Indian Institute of Technology Madras and B.Tech. in Computer Science and Business Systems at West Bengal University of Technology (GMIT campus), India. His work focuses on temporal generative modeling, geometry-aware computer vision, and controllable video synthesis, with particular interest in diffusion-based methods for talking-head generation and temporal consistency. He has served as a Research Trainee at the Variable Energy Cyclotron Centre (VECC), where he worked on pose- and landmark-conditioned video generation, benchmarking, and efficient deployment pipelines. He has contributed to research publications in journals, conference proceedings, and edited volumes, and is also associated with an Indian patent in neural network-based real-time analysis.
\end{minipage}
\\[1.5em]

\begin{minipage}[t]{\linewidth}
\vspace{0pt}
\includegraphics[width=\linewidth]{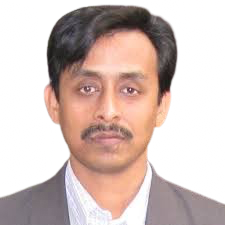}
\end{minipage}
&
\begin{minipage}[t]{\linewidth}
\vspace{0pt}
\textbf{Dr. Tapas Samanta} is a senior scientist and Head of the Computer and Informatics Group at the Variable Energy Cyclotron Centre (VECC), Department of Atomic Energy, India. With over two decades of experience, his work spans artificial intelligence, industrial automation, embedded systems, high-performance computing, and accelerator control systems. He also leads technology transfer initiatives and public scientific outreach at VECC.
\end{minipage}
\\

\end{tabular}

\end{document}